\documentclass{article}

\PassOptionsToPackage{numbers, compress}{natbib}

\usepackage[preprint]{neurips_2026}

\usepackage[utf8]{inputenc}
\usepackage[T1]{fontenc}
\usepackage{hyperref}
\usepackage{url}
\usepackage{booktabs}
\usepackage{amsfonts}
\usepackage{nicefrac}
\usepackage{microtype}
\usepackage{xcolor}
\usepackage{amsmath, amssymb}
\usepackage{algorithm}
\usepackage{algpseudocode}

\usepackage[dvipsnames]{xcolor}

\usepackage{graphicx}
\usepackage{booktabs}  
\usepackage{diagbox}
\usepackage{url}
\usepackage{multirow}
\usepackage{subcaption}
\usepackage{wrapfig}

\usepackage{xparse}
\usepackage{xspace}
\usepackage{tikz}
\usepackage{multirow}
\usepackage{colortbl}
\usepackage{array}
\newcommand{\PreserveBackslash}[1]{\let\temp=\\#1\let\\=\temp}
\newcolumntype{C}[1]{>{\PreserveBackslash\centering}p{#1}}
\newcolumntype{R}[1]{>{\PreserveBackslash\raggedleft}p{#1}}
\newcolumntype{L}[1]{>{\PreserveBackslash\raggedright}p{#1}}

\usepackage{microtype}
\usepackage{graphicx}
\usepackage{mathtools}
\usepackage{subcaption}
\usepackage{booktabs} 

\usepackage{amssymb}
\usepackage{pifont}

\usepackage{bbold}

\usepackage{amssymb,enumitem}

\setlist[itemize]{leftmargin=*}
\setlist[enumerate]{leftmargin=*}

\newcommand*{\rej}{{\ooalign{\lower.3ex\hbox{$\sqcup$}\cr\raise.4ex\hbox{$\sqcap$}}}}

\usepackage{stackengine}

\usepackage{xcolor}

\newcommand{\ours}{\texttt{NU-DUI}\xspace} 
\newcommand{\ourslong}{{Negative Unlabeled Dataset Usage Inference}\xspace}

\graphicspath{{images/}}

\usepackage{booktabs,arydshln}
\makeatletter
\def\adl@drawiv#1#2#3{%
        \hskip.5\tabcolsep
        \xleaders#3{#2.5\@tempdimb #1{1}#2.5\@tempdimb}%
                #2\z@ plus1fil minus1fil\relax
        \hskip.5\tabcolsep}
\newcommand{\cdashlinelr}[1]{%
  \noalign{\vskip\aboverulesep
           \global\let\@dashdrawstore\adl@draw
           \global\let\adl@draw\adl@drawiv}
  \cdashline{#1}
  \noalign{\global\let\adl@draw\@dashdrawstore
           \vskip\belowrulesep}}
\makeatother

\newcommand{\nlp}[1]{}

\newcolumntype{x}[1]{>{\centering\arraybackslash\hspace{0pt}}p{#1}}

\usepackage{amsmath}

\usepackage{amsmath,amsfonts,bm}

\def\eqref#1{equation~\ref{#1}}

\def\1{\bm{1}}

\DeclareMathAlphabet{\mathsfit}{\encodingdefault}{\sfdefault}{m}{sl}
\SetMathAlphabet{\mathsfit}{bold}{\encodingdefault}{\sfdefault}{bx}{n}

\title{Dataset Usage Inference without Shadow Models or Held-out Data}

\author{%
  Wojciech Łapacz\thanks{Equal contribution.}\textsuperscript{~~~}\thanks{Contact: \texttt{wojciech.lapacz.stud@pw.edu.pl}}\\
  Warsaw University of Technology\\
  \And
  Stanisław Pawlak\footnotemark[1] \\
  Warsaw University of Technology \\
  \AND
  Jan Dubiński\footnotemark[1] \\
  Warsaw University of Technology \\
  NASK National Research Institute \\
  \And
  Franziska Boenish \\
  CISPA Helmholtz Center for Information Security \\
  \And
  Adam Dziedzic \\
  CISPA Helmholtz Center for Information Security \\
}

\begin{document}

\maketitle

\begin{abstract}
How much of my data was used to train a machine learning model? \textit{Dataset Usage Inference (DUI)} aims to answer this by estimating what fraction of a dataset contributed to a model's training. However, existing DUI methods rely on assumptions that rarely hold in practice: they require training expensive shadow models to imitate the target model, and they assume access to both known training samples and an in-distribution held-out set confirmed to be absent from training. These conditions make current approaches impractical for modern large models and real data ownership disputes.
We introduce a practical DUI framework that removes these constraints. Our method requires neither shadow models nor real held-out data. Instead, it generates synthetic non-member samples, extracts diverse membership signals, and casts DUI as a mixture proportion estimation problem to estimate what share of the candidate dataset was used during training.
Experiments on large image generative models show that our method reliably quantifies dataset usage, providing a practical tool for data owners to determine how much of their data was used to train a model.

\end{abstract}

\section{Introduction}

With the rise of large generative models, the question of what data was used for training has become a major legal and ethical concern. This is no longer purely a technical issue: it has direct implications for copyright, consent, and accountability, as illustrated by ongoing disputes such as The New York Times suing OpenAI~\citep{NYT2023Complaint} or
\textit{Getty Images v. Stability AI}~\cite{getty_stability_uk_2025}.

These developments highlight a clear need for methods that can reliably check whether specific datasets were used to train a generative model. In response, researchers have explored several directions for training-data detection. One line of work watermarks potential training data or backdoors the trained model so that training leaves a detectable trace~\citep{Li2023BlackBoxWatermark,Li2020BackdoorWatermark, Tang2023CleanLabelBackdoor,Li2022UntargetedBackdoorWM}; this requires intervening before training and is therefore unavailable to most third-party auditors. A second line, Membership Inference Attacks (MIAs), instead operates post-hoc and asks whether a particular sample was part of the training set, exploiting the fact that models tend to behave differently on samples seen during training than on unseen ones~\citep{Hu2022MembershipBackdoor}. For large modern generative models, however, per-sample membership signals are typically weak and sensitive to distribution shift between member and non-member data, and most existing MIAs degrade to near-random performance on unbiased benchmarks~\citep{maini2024llmdatasetinferencedid,dubinski2025cdi}. 

Dataset Inference (DI) attacks address this weakness by aggregating membership evidence across many samples to obtain a stronger, dataset-level signal~\citep{Maini2021DatasetInference}.
Although DI methods provide useful evidence of dataset usage, they address a binary question: was the dataset used in training or not? In practice, this binary view is insufficient. Legal frameworks and real-world data ownership disputes require quantifying the \textit{extent} of usage. For example, Section~107 of the U.S. Copyright Act~\citep{USC1976FairUse} requires fair-use analysis to consider the ``amount and substantiality of the portion used'' in relation to the copyrighted work. A practical dataset audit must therefore not only detect whether a dataset was used, but also estimate what fraction of it was used.

\begin{figure}[t!]
    \centering
    \includegraphics[width=\columnwidth]{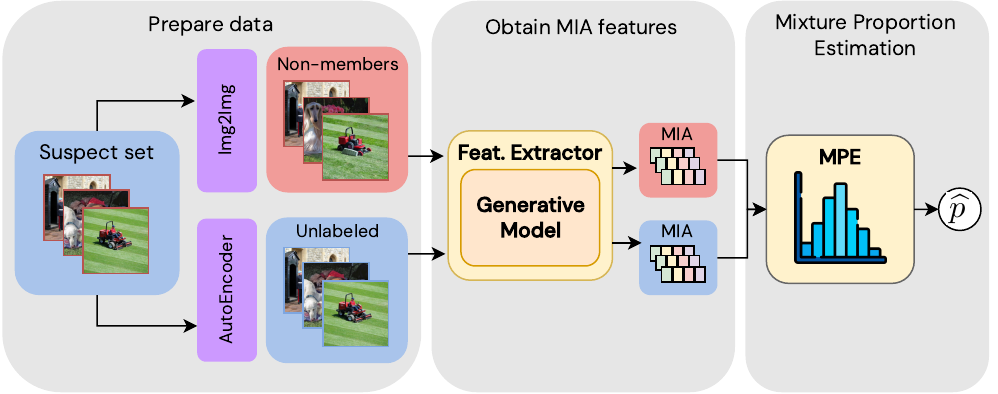}
    \caption{\textbf{Overview of \ours{}.} Given a suspect set, we generate a synthetic non-member reference set with Stable Diffusion img2img and autoencode the suspect set with the same autoencoder to reduce distribution shift. We then query the target model on both sets, extract MIA features, and apply mixture proportion estimation to infer what fraction of the suspect set was used for training.}
    \label{fig:overview}
    \vspace{-0.5cm}
\end{figure}

\citet{tongmuch} formalized this problem and proposed DUCI, which uses debiased MIAs to estimate the proportion of a dataset used during training. Their method, however, rests on two assumptions that rarely hold in realistic settings. First, it requires training expensive shadow models that closely mimic the target model's architecture and behavior; for modern large-scale generative models this is computationally prohibitive: a single shadow model for a 1.5B-parameter image autoregressive target already takes more than 1{,}500 A100 hours, and the typical five-shadow setup multiplies this further. Second, it assumes access to an in-distribution held-out set that is verified to be excluded from training; in real disputes such trusted held-out data is rarely available.

To address these limitations, we introduce \ourslong (\ours), a practical framework that needs neither shadow models nor a real held-out set. Given a suspect set, \ours{} (i) constructs a synthetic non-member reference by image-to-image paraphrasing with a generator from a \emph{different family} than the audited model, which prevents the synthetic set from sitting at a fixed point of the target's loss; (ii) autoencodes the suspect set with the same autoencoder used during paraphrasing, so that both sides of the comparison share the same generator artifacts and the residual distribution shift is dominated by membership; (iii) extracts a vector of MIA features tailored to the target model family; and (iv) applies mixture proportion estimation to recover a single estimated member ratio $\hat p$. The method requires only the dataset under examination, making it lightweight and applicable to large modern generative models.

Empirically, \ours{} produces accurate member-ratio estimates across five large-scale image generative models, both diffusion and autoregressive, without ever training a shadow model. On RAR-XXL~\cite{rar_yu2024randomizedautoregressivevisualgeneration} with a suspect set of 1{,}000 images, the full \ours{} pipeline takes about 42.5 A100 minutes, compared with the more than 1{,}500 A100 hours that would be required to execute DUCI with a single shadow model for the same target: a speedup of more than $2{,}000$ times. With the autoencoded suspect set, the mean estimation error stays well below 0.1 across most target models (e.g., $\text{MAE}=0.058$ on RAR-XXL with PUL-based MPE), closing most of the gap to the oracle setting that requires a trusted held-out non-member set. 

\textbf{In summary, our main contributions are:}
\begin{itemize}[itemsep=2pt, parsep=0pt, topsep=0pt]
    \item We recast Dataset Usage Inference as a Mixture Proportion Estimation problem, removing the need for expensive shadow models.
    \item We propose generating realistic synthetic non-member samples and autoencoding the suspect set, eliminating the reliance on a trusted real held-out set, which is rarely available in practice.
    \item We empirically demonstrate that \ours{} produces accurate member-ratio estimates on five large-scale autoregressive and diffusion image generators at orders-of-magnitude lower compute than shadow-model-based DUCI, unlocking dataset usage inference for modern generative models.
\end{itemize}

\section{Background and Related Work}

This section reviews the three lines of work \ours{} builds on: membership inference attacks, dataset inference, and mixture proportion estimation.

\paragraph{Membership Inference Attacks.}
Membership Inference Attacks (MIAs) aim to determine whether a specific data point was included in a model's training set. They typically exploit overfitting: models tend to behave systematically differently on samples they have seen during training compared to unseen samples.

In large language models (LLMs), MIAs commonly rely on likelihood-based signals, such as token-level losses~\citep{yeom2018privacy}, unusually confident token predictions~\citep{shi2024detecting}, or combinations of likelihood-ratio tests~\citep{zarifzadeh2023low}. \citet{kowalczuk2025privacy} extend these ideas to image autoregressive models (IARs), showing that attacks developed for LLMs transfer to autoregressive image generators. For diffusion models, MIAs typically estimate membership through noise-prediction errors at selected timesteps. SecMI~\citep{duan23bSecMI} compares errors between sampling and inverse-sampling trajectories, PIA~\citep{kong2024an} contrasts predictions on clean and noised inputs, and CLiD~\citep{clid} measures membership via discrepancies in conditional likelihoods.

A recurring finding, however, is that MIAs are highly sensitive to distribution shifts between member and non-member data. On unbiased benchmarks for large modern models, most existing attacks degrade substantially and often perform only marginally above random guessing~\citep{maini2024llmdatasetinferencedid, dubinski2025cdi}. This motivates aggregating membership evidence across many samples rather than relying on per-sample decisions.

\paragraph{Dataset Inference.}
Dataset Inference (DI)~\citep{Maini2021DatasetInference} aims to determine whether a specific dataset was included in a model's training set. Unlike MIAs, which operate on individual samples, DI aggregates membership signals across many points to obtain a dataset-level decision. Formally, given a target model $f$ and a suspect dataset $D_{\text{sus}}$, DI tests whether $f$ was trained on $D_{\text{sus}}$ by comparing its membership signals on $D_{\text{sus}}$ with those on a held-out dataset $D_{\text{val}}$ drawn from the same distribution.

The general DI pipeline consists of three steps: (1) extract membership features for samples in both $D_{\text{sus}}$ and $D_{\text{val}}$; (2) aggregate these features into a dataset-level score; and (3) apply a statistical test to compare the scores between the two sets. The choice of membership features depends on the learning paradigm, ranging from decision-boundary signals in supervised models~\citep{Maini2021DatasetInference} to representational statistics in self-supervised settings~\citep{dziedzic2022dataset}. Modern DI methods for generative models, including LLMs~\citep{maini2025reassessing}, diffusion models~\citep{dubinski2025cdi}, and image autoregressive models~\citep{kowalczuk2025privacy}, all follow this template: they derive membership features using MIAs designed for the specific model family and aggregate them into a dataset-level indicator of training usage. DI, however, still answers a binary question; it does not quantify what fraction of $D_{\text{sus}}$ was actually used.

\paragraph{Mixture Proportion Estimation.}
Mixture Proportion Estimation (MPE) is the task of identifying the fraction of positive instances, $\alpha$, within an unlabeled dataset, a problem central to Positive--Unlabeled Learning (PUL)~\citep{liu2002partially}. Early theoretical foundations by \citet{blanchard2010noveltydetection} and \citet{scott2015ARO} established the \textit{irreducibility condition} required for identifiability, but their initial estimators were often computationally infeasible. Practical estimators followed from \citet{elkan2008learning}, who proposed leveraging a positive-vs-unlabeled classifier, and \citet{plessis2015convexpul}, who used Pearson divergence minimization; both, however, typically rely on the restrictive assumption of disjoint support between positive and negative distributions. \citet{ramaswamy2016km} introduced the KM algorithm, providing the first computationally feasible approach with convergence guarantees by embedding distributions into a Reproducing Kernel Hilbert Space, although it scales poorly in high-dimensional settings. Tree-based methods such as TIcE~\citep{bekker2018tice} partition the feature space to identify regions where positives occur at a higher rate than in the overall mixture but can become unreliable on complex features. Classifier-based heuristics such as AlphaMax~\citep{jain2016alphamax} and DEDPUL~\citep{ivanov2020dedpul} treat a positive-vs-unlabeled classifier as a dimensionality-reduction tool and estimate $\alpha$ from the resulting one-dimensional score; they are efficient but tend to depend strongly on classifier quality. MPE can also be framed as histogram-based estimation, where the outputs of a trained PU classifier~\citep{bekker2019sarem,gong2021lbe,teser2025threshold} are binned to estimate the positive-class likelihood within the unlabeled mixture. Our work uses MPE as a modular tool, leveraging the fact that, under the ``selected completely at random'' (SCAR) assumption, MPE provides a lower bound on the class prior.

\definecolor{lightblue}{RGB}{31,119,180}

\section{Method}
\label{sec:method}

We estimate the fraction $p$ of a suspect set $X^{\text{real}}$ that was used to train a target generative model $M$. The suspect set may contain both members and non-members, and we assume access only to $X^{\text{real}}$ and to $M$ at inference time: no shadow models and no trusted held-out non-member set. Our pipeline (Algorithm~\ref{alg:global}) has three stages: (i) we construct a synthetic non-member reference set $X^{\text{img2img}}_{\text{nonmember}}$ together with an autoencoded version $X^{\text{AE}}$ of the suspect set, both designed to limit the distribution shift between member-side and non-member-side data; (ii) we extract membership-inference (MIA) features from both sets using attacks tailored to the target model family; and (iii) we apply mixture proportion estimation (MPE) to the resulting features to estimate $p$.

\begin{figure}[h!]
\centering
\begin{minipage}{0.8\linewidth}
\hrule height 1.5pt
\vspace{2pt}
\captionof{algorithm}{\textit{\ourslong{}}}
\label{alg:global}
\vspace{2pt}
\hrule
\begin{algorithmic}[1]
\small
\Require Suspect set $X^{\text{real}}$; feature extractor $\Phi_{\text{MIA}}$

\State Construct $X^{\text{img2img}}_{\text{nonmember}}$ by image-to-image paraphrasing of $X^{\text{real}}$
\State Construct $X^{\text{AE}}$ by autoencoding $X^{\text{real}}$
\State $U \gets X^{\text{AE}}$ \Comment{unlabeled}
\State $N \gets X^{\text{img2img}}_{\text{nonmember}}$ \Comment{synthetic negatives}
\State
\[
Z^{\text{MIA}} = \Phi_{\text{MIA}}(U \cup N)
\]
\State $\hat p \gets \textsc{MPE}(Z^{\text{MIA}}, U, N)$
\State \Return $\hat p$

\end{algorithmic}
\vspace{2pt}
\hrule height 1.5pt
\end{minipage}
\vspace{-0.4cm}
\end{figure}

\paragraph{Synthetic Held-Out Set Generation}\label{subsec:synt_hold_out_generation}

A central obstacle in dataset usage inference is the absence of a trusted held-out set known to contain only non-members~\citep{zhao2025posthocDI}. Prior dataset inference and dataset usage inference methods sidestep this by assuming access to such a set, but in real auditing scenarios it is typically unavailable. We instead build our reference data directly from the suspect set $X^{\text{real}}$: a synthetic non-member set $X^{\text{img2img}}_{\text{nonmember}}$, generated by paraphrasing the suspect images, and an autoencoded suspect set $X^{\text{AE}}$, obtained by encoding and decoding $X^{\text{real}}$ with the same autoencoder used for paraphrasing. We use $X^{\text{AE}}$ rather than $X^{\text{real}}$ as the unlabeled set so that both sides of the comparison share the same generative-model artifacts, reducing the distribution shift between the unlabeled mixture and the synthetic negatives.

A key design choice is to generate synthetic non-members with a model from a \emph{different family} than the audited model. For image autoregressive target models we use Stable Diffusion~\citep{rombach2022high}; for diffusion target models we use a VAR model~\citep{var_tian2024visualautoregressivemodelingscalable}. The reason is that paraphrasing an image with a model from the same family produces samples that lie near a fixed point of that family's loss: e.g., a diffusion-paraphrased image is, by construction, a preferred state of the diffusion reverse process and exhibits unusually low denoising error. If such samples were used as synthetic non-members against a diffusion target, they would have artificially low diffusion loss and would be trivially separable from suspect samples for reasons unrelated to membership. Cross-family generation breaks this confound, because different generative families minimize different losses.

\textit{Synthetic non-members.}
To build the synthetic non-member set, we apply an image-to-image generator $\mathcal{G}$ to each suspect image $x^{\text{real}}$, conditioned on a class-level prompt $p_c$:
\[
    x^{\text{img2img}}_{\text{nonmember}}
    = \mathcal{G}\!\left( x^{\text{real}},\, p_c;\, s, g \right),
\]
where $s$ controls the noise injected into the input image and $g$ is the guidance scale. Compared with unconditional generation from noise, image-to-image paraphrasing better preserves coarse semantic content and low-frequency structure, which is important for keeping the synthetic set distributionally close to the suspect data. Because these outputs are newly generated, we treat them as non-members and use them as the negative reference set $N$.\footnote{We adopt the set-theoretic definition of non-membership from~\citet{yeom2018privacy}: $x$ is a non-member of $M$ iff $x \notin \mathcal{D}_{\text{train}}(M)$. While paraphraser memorization can in principle produce a near-verbatim copy of a training image, such cases are documented to occur at rates orders of magnitude too low to influence aggregate MPE estimates at our suspect-set sizes~\citep{carlini2023extracting_diffusion}. Our aim is the best practical solution under realistic auditing constraints, which we validate empirically in Section~\ref{sec:empirical_evaluation}.}

\begin{figure}[h!]
    \centering
    \includegraphics[width=\linewidth, trim=0cm 0cm 0cm 6cm, clip]{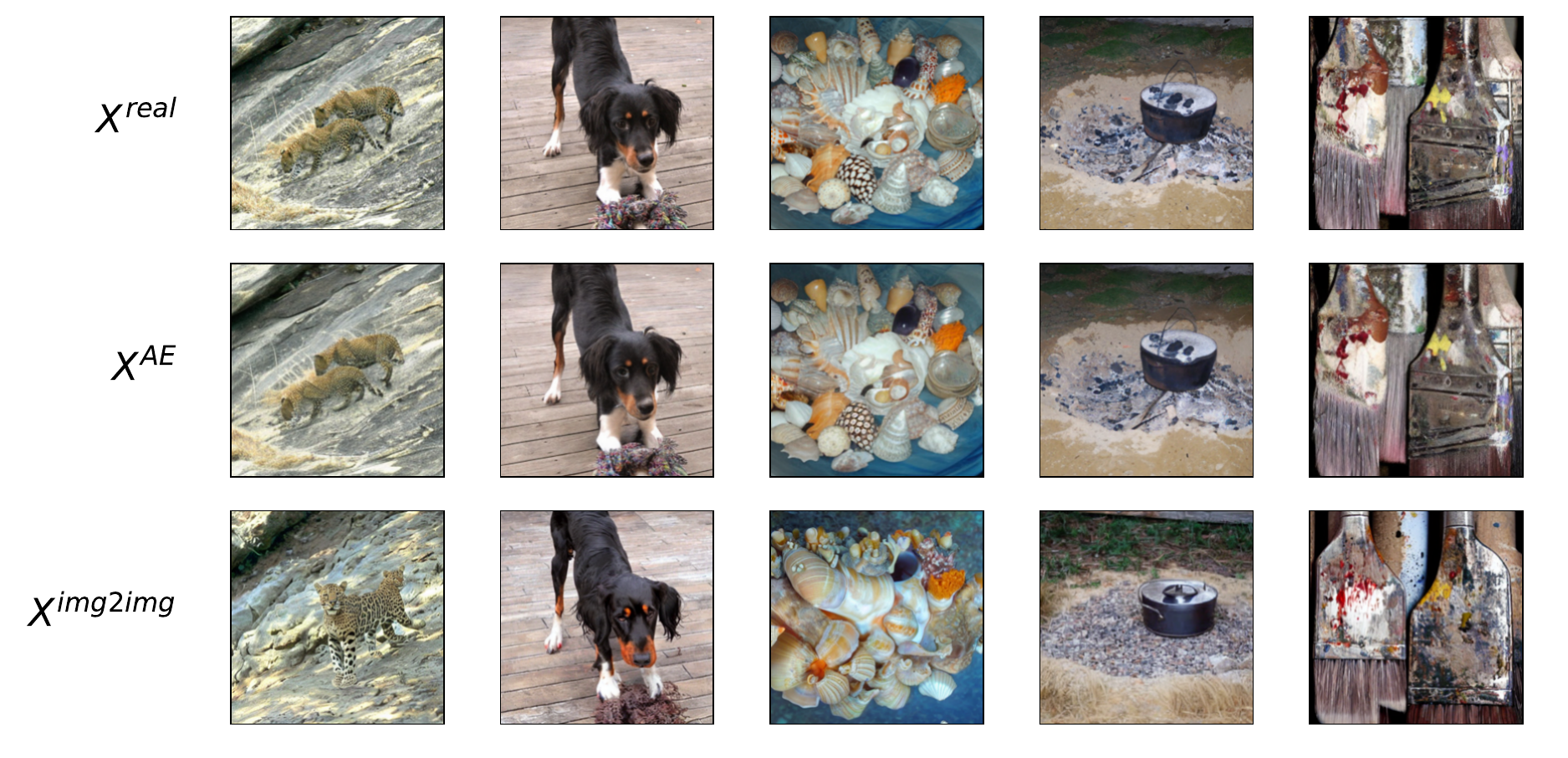}
    \caption{\textbf{Examples of synthetic non-members and autoencoded suspect images.} For each real ImageNet sample $x \in X^{\text{real}}$, we show the autoencoded reconstruction $x^{\text{AE}} = \mathcal{D}(\mathcal{E}(x))$ that we use as the unlabeled set $U$, and the cross-family image-to-image paraphrase $x^{\text{img2img}}_{\text{nonmember}} = \mathcal{G}(x, p_c;\, s, g)$ that we use as the synthetic non-member set $N$. }
    \label{fig:synth_strength_examples}
\end{figure}

\textit{Autoencoded suspect set.}
Images produced by generative models have been shown to differ from real images in the high-frequency domain~\citep{frequency}. This gap is fundamental and difficult to remove from the synthetic side. Rather than attempting to remove this artifact from the synthetic side, we introduce the \emph{same} shift on the suspect side, by reconstructing each suspect image with the autoencoder used during generation:
\[
    z = \mathcal{E}\!\left( x^{\text{real}} \right), \qquad
    \tilde{x}^{\text{AE}} = \mathcal{D}(z).
\]
This preserves the semantic content and label of each image while injecting generator-specific high-frequency artifacts comparable to those present in $X^{\text{img2img}}_{\text{nonmember}}$. We denote the resulting set by $X^{\text{AE}}$ and use it as the unlabeled set $U$. Figure~\ref{fig:synth_strength_examples} shows examples of the resulting samples; Appendix~\ref{app:construct_synth} provides details for the full construction.

\vspace{0.2cm}

\paragraph{Extracting MIA Features}
\label{subsec:mia_features}
Given the unlabeled set $U$ and the synthetic negative set $N$, we extract a feature vector $\Phi_{\text{MIA}}(x) \in \mathbb{R}^{d_M}$ for each sample using a suite of MIAs tailored to the target model family:
\[
\Phi_{\text{MIA}}(x) = \left[f_1(x), \ldots, f_{d_M}(x)\right],
\]
where each $f_m(x)$ is a scalar membership score obtained by querying the target model on $x$.

We use multiple MIA signals rather than a single one because no single signal is reliably effective on large modern generative models: per-sample membership cues are typically weak and noisy, and combining complementary signals into a single feature vector yields better separation between member and non-member distributions~\citep{maini2024llmdatasetinferencedid, dubinski2025cdi, kowalczuk2025privacy}. For image autoregressive target models we use the feature suite of \citet{kowalczuk2025privacy}; for diffusion target models we use that of \citet{dubinski2025cdi}. Beyond this choice, our method is conceptually agnostic to the specific MIA features, as long as they provide some separation between member and non-member score distributions.

\paragraph{Estimating the Member Ratio $p$}
\label{subsec:MPE_method}
We cast dataset usage inference as a mixture proportion estimation problem. After constructing $U = X^{\text{AE}}$ and $N = X^{\text{img2img}}_{\text{nonmember}}$ and computing the MIA feature vectors for both sets, we model the unlabeled score distribution as a two-component mixture:
\[
    p_U(s) = \pi\, p_M(s) + (1-\pi)\, p_N(s),
\]
where $p_M(s)$ and $p_N(s)$ are the score distributions of members and non-members, and $\pi$ is the unknown member proportion in $U$.

We then run a standard MPE estimator on the MIA feature vectors, treating the synthetic samples in $N$ as labeled non-members and the autoencoded suspect samples in $U$ as the unlabeled mixture. The estimator returns $\hat p$, our estimate of $\pi$, i.e., the predicted fraction of the suspect set used during training. The framework is agnostic to the choice of MPE estimator; we evaluate several standard ones in Section~\ref{sec:empirical_evaluation}.

\section{Empirical Evaluation}
\label{sec:empirical_evaluation}

\subsection{Experimental Setup}

\paragraph{Models and data.}
We evaluate \ours{} on large, state-of-the-art image autoregressive models VAR-24 (1.0B) and VAR-30 (2.1B)~\citep{var_tian2024visualautoregressivemodelingscalable}, RAR-XL (955M) and RAR-XXL (1.5B)~\citep{rar_yu2024randomizedautoregressivevisualgeneration}, and the diffusion model DiT-RF-XL/2-8E2A (4.1B total, 1.5B active parameters)~\citep{fei2024scalingdiffusiontransformers16}, all trained for class-conditioned generation on ImageNet-1k~\citep{deng2009imagenet}. To ensure scientifically sound evaluation, we restrict ourselves to models trained on public datasets with well-defined IID train/test splits~\citep{dubinski2024towards}. The same restriction applies to prior MIA and DI works, since without an IID split, observed differences between member and non-member sets cannot be reliably attributed to training membership. Table~\ref{tab:iar_model_details} summarizes the architecture, training budget, and reported FID-50K of each evaluated model.

\begin{table}[h!]
    \centering
    \scriptsize
    \caption{\textbf{Model details for the evaluated target models.} For each image autoregressive (VAR-\textit{d}24, VAR-\textit{d}30, RAR-XL, RAR-XXL) and diffusion (DiT-RF-XL/2-8E2A) target, we report total and active parameter counts, the training budget (epochs or iterations), and the reported FID-50K on ImageNet-1k.}
    \vspace{0.1cm}
    \label{tab:iar_model_details}
    \begin{tabular}{l c c c c c}
        \toprule
        & \multicolumn{2}{c}{\textbf{VAR Models}} & \multicolumn{2}{c}{\textbf{RAR Models}}
        & \textbf{DiT-RF}\\
        \cmidrule(r){2-3} \cmidrule(r){4-5} \cmidrule(r){6-6}
        & VAR-\textit{d}24 & VAR-\textit{d}30 & RAR-XL & RAR-XXL & DiT-RF-XL/2-8E2A \\
        \midrule
        \textbf{Total parameters} & 1.0B & 2.1B & 955M & 1.5B & 4.1B \\
        \textbf{Active parameters} & 1.0B & 2.1B & 955M & 1.5B & 1.5B \\
        \textbf{Training steps} & 500k & 438k & 250k  & 250k & 7M \\
        \textbf{FID-50K} & 2.33 & 1.92 & 1.50 & 1.48 & 1.72 \\
        \bottomrule
    \end{tabular}
\end{table}

\paragraph{Evaluation procedure.}
Our evaluation simulates a realistic auditing scenario: a user is interested in whether specific data points were part of the model's training set, and to what extent. Given a suspect set $U$, \ours{} produces a single estimated member ratio $\hat p$. We compare three reference conditions for the negative set $N$:

\begin{itemize}
    \item \textbf{Real non-members.} An idealized but unrealistic setting in which $N$ contains genuine held-out non-members from the same distribution as the non-member portion of $U$. This assumes access to a trusted held-out set known to be excluded from training, which is rarely available in practice. Because the non-member reference matches the test-time distribution, this setting yields the lowest achievable estimation error.
    \item \textbf{Synthetic non-members.} A synthetic $N$ obtained by applying our image-to-image paraphrasing pipeline to samples from $U$, used directly together with the unmodified suspect set as the unlabeled mixture. Because the synthetic samples carry generation artifacts that the real suspect samples do not, the MPE estimator can spuriously separate the two sets based on those artifacts rather than on membership cues, leading to inflated estimation errors.
    \item \textbf{Synth + AE (\ours{}).} The full \ours{} pipeline: the same synthetic $N$ as above, but with the suspect set additionally autoencoded so that both sides of the comparison share the same generator-induced artifacts. This shifts the estimator back toward true membership signals.
\end{itemize}

We vary the true member ratio $p \in (0, 1]$ and the suspect-set size $|U|$ to assess robustness across realistic auditing conditions. We exclude $p = 0$ because in that case the suspect set contains no members at all, which reduces to the binary dataset inference question (was $D_{\text{sus}}$ used to train $f$ or not?) and is already handled by DI methods for image autoregressive~\citep{kowalczuk2025privacy} and diffusion~\citep{dubinski2025cdi} models.

\subsection{Results}

\begin{figure*}[t!]
    \centering

    \begin{subfigure}[t]{0.31\textwidth}
        \vspace{0pt}
        \centering
        \includegraphics[width=\linewidth]{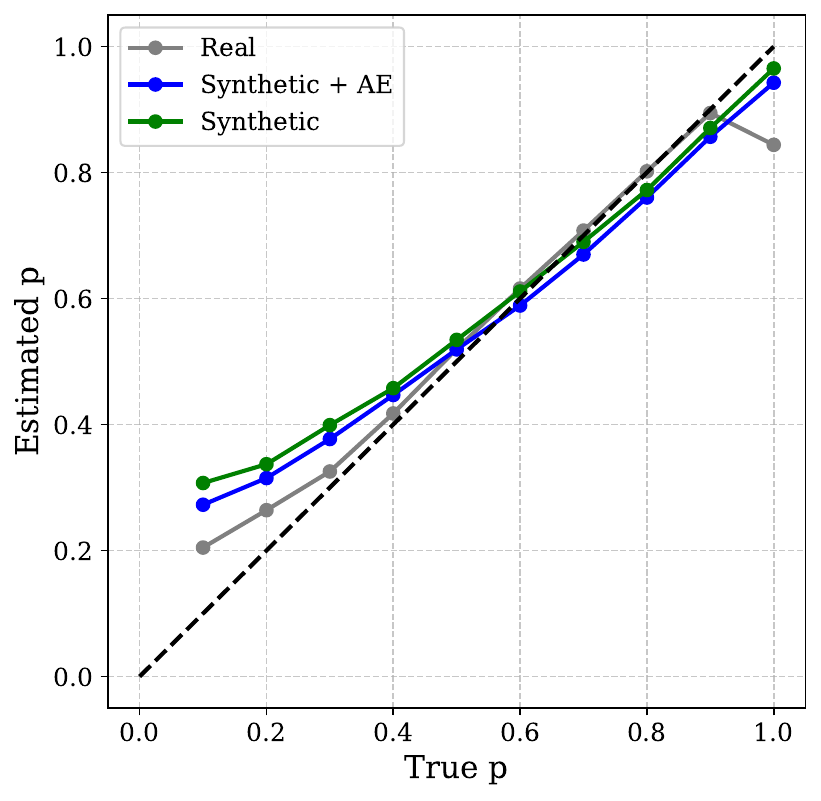}
        \caption{\texttt{RAR-XXL}}
    \end{subfigure}
    \hfill
    \begin{subfigure}[t]{0.31\textwidth}
        \vspace{0pt}
        \centering
        \includegraphics[width=\linewidth]{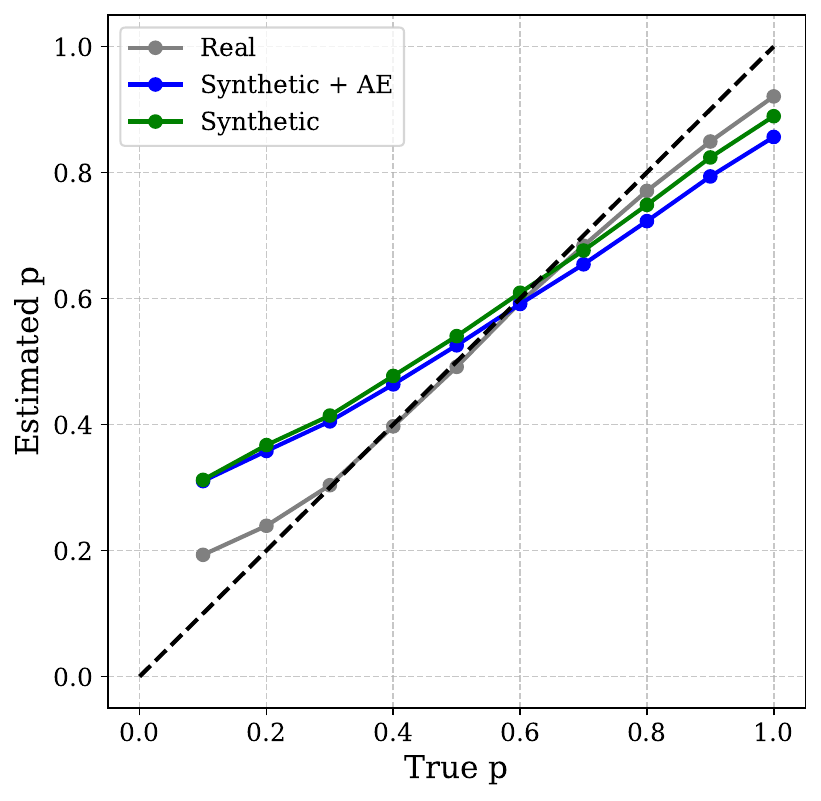}
        \caption{\texttt{RAR-XL}}
    \end{subfigure}
    \hfill
    \begin{subfigure}[t]{0.31\textwidth}
        \vspace{0pt}
        \centering
        \includegraphics[width=\linewidth]{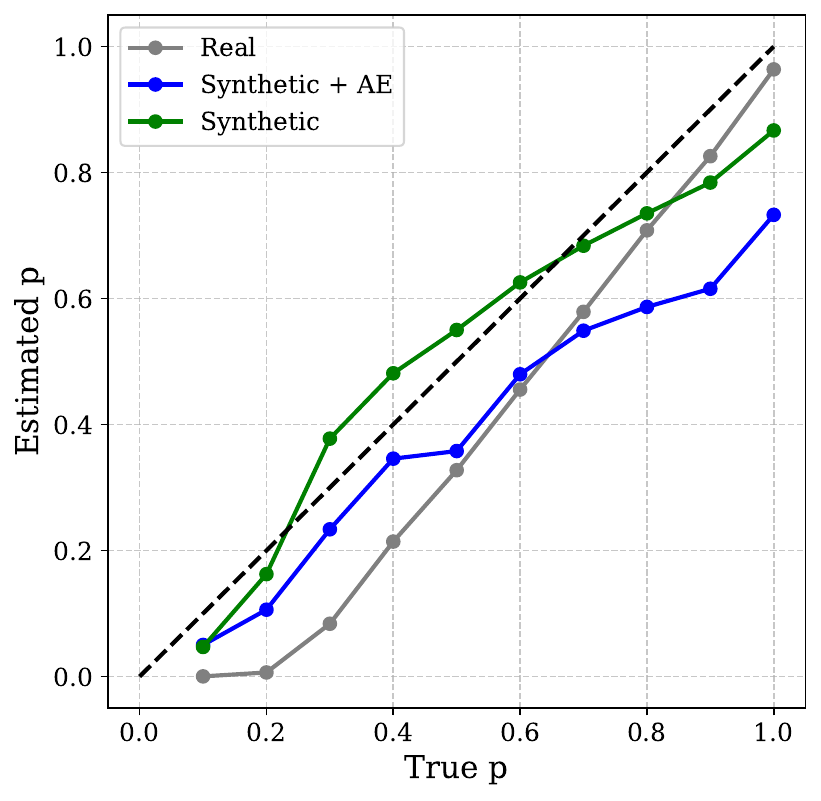}
        \caption{\texttt{DiT-RF-XL}}
    \end{subfigure}

    \vspace{1.5mm}

    \begin{subfigure}[t]{0.31\textwidth}
        \vspace{0pt}
        \centering
        \includegraphics[width=\linewidth]{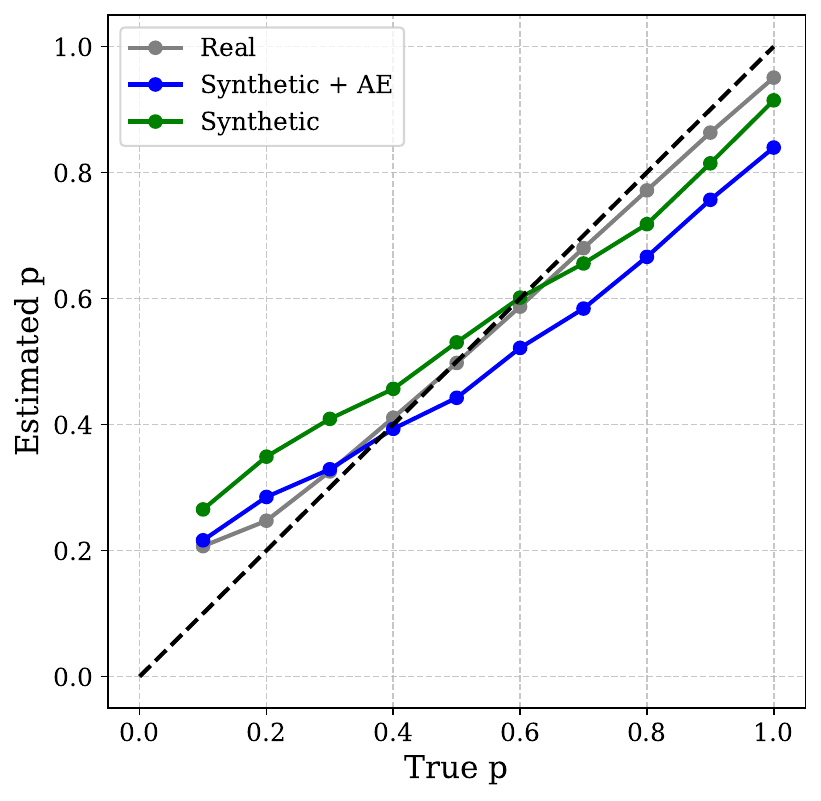}
        \caption{\texttt{VAR-30}}
    \end{subfigure}
    \hfill
    \begin{subfigure}[t]{0.31\textwidth}
        \vspace{0pt}
        \centering
        \includegraphics[width=\linewidth]{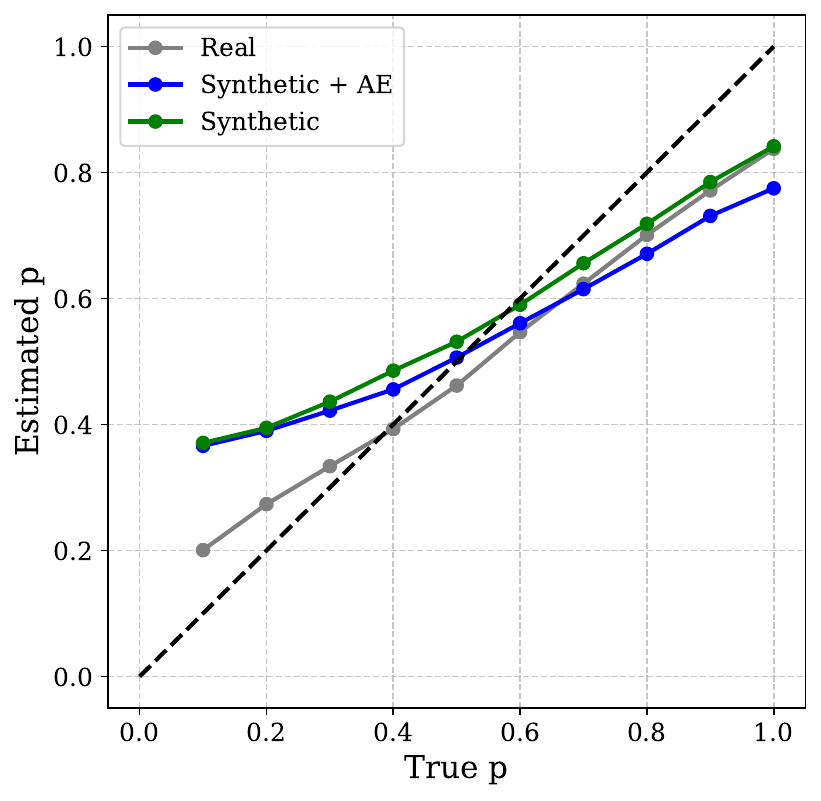}
        \caption{\texttt{VAR-24}}
    \end{subfigure}
    \hfill
    \begin{minipage}[t]{0.31\textwidth}
        \vspace{0pt}
        \captionsetup{type=figure}
        \caption{\textbf{Estimated member ratios.} We show estimated member ratios across evaluated models for varying ground-truth $p$ under three reference conditions: \textit{real non-members} (ideal but unrealistic), \textit{synthetic non-members} with real unlabeled set, and \textit{synthetic non-members with autoencoded unlabeled set} (our \ours{}).}
        \label{fig:dui_results}
    \end{minipage}
\vspace{-0.4cm}
\end{figure*}

\paragraph{Accurate member-ratio estimation across models.}
Figure~\ref{fig:dui_results} visualizes the estimated member ratios as a function of the ground-truth $p$, and Table~\ref{tab:main} reports the corresponding MAE and max MAE across all tested $p$. In the Real setting, \ours{} stays close to the ground-truth ratio across all evaluated autoregressive and diffusion target models, with the estimated curves tracking the identity line. This setting also yields the lowest estimation errors in Table~\ref{tab:main}, confirming that the MIA features used by \ours{} are sufficiently discriminative when the non-member reference matches the test-time distribution. When a confirmed held-out non-member set is available, \ours{} reduces to exactly this Real setting; therefore \textbf{\textit{Real} is not only an oracle baseline, but also the performance achievable by \ours{} when held-out non-member data is accessible.} Table~\ref{tab:main} also shows that the result is not specific to a single MPE estimator: TIcE, AlphaMax, and the two PUL variants (LBE and NTC-$\tau$MI) all exhibit the same qualitative behavior in the Real regime.

\begin{table*}[b!]
\centering
\vspace{-0.3cm}
\small
\renewcommand{\arraystretch}{1}
\setlength{\tabcolsep}{2pt}
\vspace{0.1cm}
\caption{\textbf{Dataset usage estimation under different MPE methods} (suspect set size 1{,}000 images). For each target model and reference condition (Real, Synth, Synth + AE), we report the \textit{MAE} and \textit{max MAE} of the estimated member ratio $\hat p$, taken as the mean and maximum absolute error across all tested ground-truth $p \in (0, 1]$. Across all four MPE methods, the \textit{Synth} reference systematically degrades estimation relative to the \textit{Real} oracle, and \textit{Synth + AE} recovers most of this gap on the autoregressive targets, confirming that autoencoding the suspect set is the key ingredient that makes synthetic-only references usable in practice. Values are reported with standard deviation across 3 runs; lower is better.}
\label{tab:main}
\resizebox{\textwidth}{!}{%
\begin{tabular}{@{}c@{}|lcccccccccc}
\toprule
Data & \textbf{Method} & \multicolumn{2}{c}{RAR-XL} & \multicolumn{2}{c}{RAR-XXL} & \multicolumn{2}{c}{VAR-24} & \multicolumn{2}{c}{VAR-30} & \multicolumn{2}{c}{DiT-RF-XL} \\
\cmidrule(lr){3-4} \cmidrule(lr){5-6} \cmidrule(lr){7-8} \cmidrule(lr){9-10} \cmidrule(lr){11-12}
& & \textbf{MAE} & \textbf{max MAE} & \textbf{MAE} & \textbf{max MAE} & \textbf{MAE} & \textbf{max MAE} & \textbf{MAE} & \textbf{max MAE} & \textbf{MAE} & \textbf{max MAE}\\
\midrule
\multirow{4}{*}{\rotatebox{90}{Real}} & TIcE~\citep{bekker2018tice} & 0.110{\tiny $\pm$0.03} & 0.142 & 0.086{\tiny $\pm$0.02} & 0.112 & 0.174{\tiny $\pm$0.04} & 0.217 & 0.097{\tiny $\pm$0.02} & 0.117 & 0.095{\tiny $\pm$0.03} & 0.140 \\
& AlphaMax~\citep{jain2016alphamax} & 0.060{\tiny $\pm$0.03} & 0.115 & 0.041{\tiny $\pm$0.02} & 0.073 & 0.081{\tiny $\pm$0.02} & 0.113 & 0.047{\tiny $\pm$0.03} & 0.123 & 0.065{\tiny $\pm$0.03} & 0.124 \\
& PUL (LBE~\citep{gong2021lbe}) & 0.059{\tiny $\pm$0.04} & 0.146 & 0.050{\tiny $\pm$0.06} & 0.166 & 0.085{\tiny $\pm$0.04} & 0.159 & 0.045{\tiny $\pm$0.04} & 0.134 & 0.091{\tiny $\pm$0.05} & 0.169 \\
& PUL (NTC-$\tau$MI~\citep{teser2025threshold}) & 0.045{\tiny $\pm$0.03} & 0.088 & 0.059{\tiny $\pm$0.06} & 0.200 & 0.080{\tiny $\pm$0.04} & 0.150 & 0.036{\tiny $\pm$0.03} & 0.098 & 0.066{\tiny $\pm$0.07} & 0.263 \\
\midrule
\multirow{4}{*}{\rotatebox{90}{Synth}} & TIcE~\citep{bekker2018tice} & 0.088{\tiny $\pm$0.04} & 0.136 & 0.082{\tiny $\pm$0.03} & 0.110 & 0.103{\tiny $\pm$0.06} & 0.209 & 0.059{\tiny $\pm$0.03} & 0.119 & 0.150{\tiny $\pm$0.05} & 0.210 \\
& AlphaMax~\citep{jain2016alphamax} & 0.083{\tiny $\pm$0.06} & 0.209 & 0.083{\tiny $\pm$0.06} & 0.207 & 0.117{\tiny $\pm$0.09} & 0.267 & 0.091{\tiny $\pm$0.07} & 0.257 & 0.182{\tiny $\pm$0.08} & 0.288 \\
& PUL (LBE~\citep{gong2021lbe}) & 0.104{\tiny $\pm$0.08} & 0.279 & 0.072{\tiny $\pm$0.08} & 0.245 & 0.118{\tiny $\pm$0.09} & 0.290 & 0.091{\tiny $\pm$0.06} & 0.213 & 0.249{\tiny $\pm$0.09} & 0.369 \\
& PUL (NTC-$\tau$MI~\citep{teser2025threshold}) & 0.094{\tiny $\pm$0.07} & 0.229 & 0.061{\tiny $\pm$0.05} & 0.169 & 0.132{\tiny $\pm$0.10} & 0.321 & 0.097{\tiny $\pm$0.08} & 0.248 & 0.130{\tiny $\pm$0.08} & 0.272 \\
\midrule
\multirow{4}{*}{\rotatebox{90}{Synth + AE}} & TIcE~\citep{bekker2018tice} & 0.104{\tiny $\pm$0.05} & 0.162 & 0.083{\tiny $\pm$0.03} & 0.114 & 0.118{\tiny $\pm$0.06} & 0.232 & 0.098{\tiny $\pm$0.05} & 0.191 & 0.223{\tiny $\pm$0.08} & 0.304 \\
& AlphaMax~\citep{jain2016alphamax} & 0.081{\tiny $\pm$0.05} & 0.174 & 0.081{\tiny $\pm$0.05} & 0.194 & 0.126{\tiny $\pm$0.09} & 0.315 & 0.099{\tiny $\pm$0.06} & 0.233 & 0.279{\tiny $\pm$0.16} & 0.517 \\
& PUL (LBE~\citep{gong2021lbe}) & 0.108{\tiny $\pm$0.08} & 0.269 & 0.078{\tiny $\pm$0.08} & 0.262 & 0.126{\tiny $\pm$0.08} & 0.251 & 0.101{\tiny $\pm$0.05} & 0.173 & 0.270{\tiny $\pm$0.11} & 0.409 \\
& PUL (NTC-$\tau$MI~\citep{teser2025threshold}) & 0.096{\tiny $\pm$0.06} & 0.209 & 0.058{\tiny $\pm$0.04} & 0.146 & 0.112{\tiny $\pm$0.07} & 0.220 & 0.097{\tiny $\pm$0.05} & 0.208 & 0.219{\tiny $\pm$0.13} & 0.419 \\
\bottomrule
\end{tabular}%
}
\end{table*}

\paragraph{Synthetic non-members introduce bias, which Synth + AE largely corrects.}
When the non-member reference $N$ contains only synthetic samples, the ``Synthetic'' curves in Figure~\ref{fig:dui_results} systematically overestimate the true member ratio at small $p$. This effect arises because the unmodified suspect set and the synthetic non-members differ in low-level generation artifacts, which the MPE estimator can exploit instead of using genuine membership cues. Table~\ref{tab:main} shows the same pattern: both MAE and max MAE in the Synth rows generally increase relative to Real. Autoencoding the suspect set (Synth + AE) injects matching artifacts on the suspect side, which brings the estimated curves visibly closer to the diagonal and narrows the gap to Real. This enables accurate dataset-usage estimation without any access to a trusted held-out non-member set.

\paragraph{Computational cost.}
A central practical advantage of \ours{} over shadow-model-based approaches such as DUCI is its compute footprint. Table~\ref{tab:cost} compares the estimated A100 time of \ours{} and DUCI on RAR-XXL (1.5B) for a suspect set of 1000 images. DUCI is dominated by shadow-model training: a single shadow model already requires more than 1500 hours, and the typical five-shadow setup exceeds 7500 hours, equivalent to several months of continuous A100 use. In contrast, \ours{} avoids shadow training entirely; the only costs are synthetic generation, autoencoding, target-model queries, and lightweight MPE, totaling roughly 42.5 minutes, more than $2{,}000\times$ faster than the single-shadow DUCI configuration. Beyond the raw cost, \ours{} also avoids the need to reproduce the training pipeline of the inspected model, which for modern large-scale generative models typically requires access to proprietary training data, training code, and full hardware budgets, none of which are realistic for an external auditor.

\begin{table}[h!]
\centering
\small
\renewcommand{\arraystretch}{0.9}
\setlength{\tabcolsep}{1pt}
\caption{\textbf{Estimated time comparison between \ours{} and DUCI.} For each method we report the per-stage compute cost on the target model RAR-XXL (1.5B) with a suspect set of 1{,}000 images on an Nvidia A100 GPU. The DUCI cost is dominated by shadow-model training (roughly 1{,}500 hours per shadow model). \ours{} avoids shadow training entirely, totalling 42.5 minutes, more than $2{,}000\times$ faster than the single-shadow model DUCI configuration on the same target.}
\vspace{0.1cm}
\label{tab:cost}
\begin{tabular}{lccccc|c}
\toprule
\textbf{Method} & Shadow models & SD & AE & Target & Estimation & Total \\
\midrule
DUCI~\citep{tongmuch} (1 shadow)
  & 1500 h
  & 0
  & 0
  & 16.6 min
  & $<$1 s
  & \textbf{1500.2 h} \\
DUCI~\citep{tongmuch} (5 shadows)
  & 7500.0 h
  & 0
  & 0
  & 49.8 min
  & $<$1 s
  & \textbf{7501 h} \\
\textbf{NU-DUI (ours)}
  & 0
  & 16.6 min
  & 16.6 min
  & 8.3 min
  & $<$ 1 min
  & \textbf{42.5 min} \\
\bottomrule
\end{tabular}
\end{table}

\begin{figure}[b!]
    \centering
    \begin{subfigure}{0.38\linewidth}
        \centering
        \includegraphics[width=\linewidth]{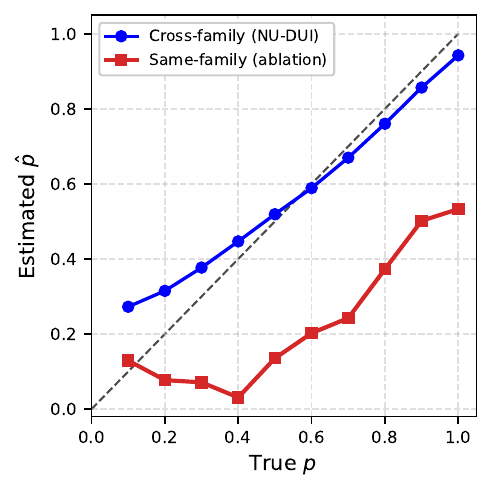}
        \caption{IAR model (\texttt{RAR-XXL})}
    \end{subfigure}
    \hspace{0.12\linewidth}
    \begin{subfigure}{0.38\linewidth}
        \centering
        \includegraphics[width=\linewidth]{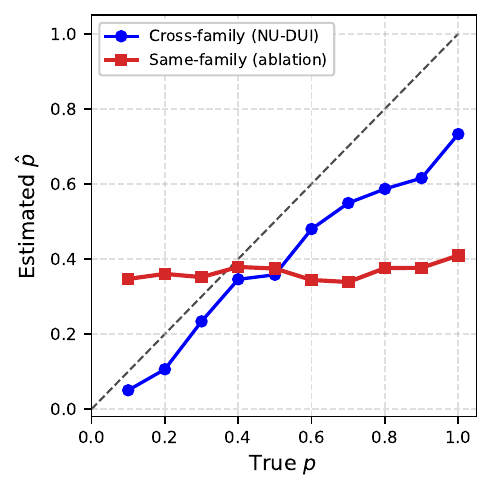}
        \caption{Diffusion model (\texttt{DiT-RF})}
    \end{subfigure}

    \caption{\textbf{Same-family vs.\ cross-family synthetic non-member generation.} Estimated $\hat p$ vs.\ true $p$. Same-family generation pulls $\hat p$ off the ground truth, while cross-family generation tracks it closely.}
    \label{fig:samefam_ablation}
\end{figure}

\paragraph{Cross-family generation supports more accurate estimation.}
A natural question is whether constructing the synthetic non-member set in Section~\ref{subsec:synt_hold_out_generation} via image-to-image paraphrasing with a generator $\mathcal{G}$ from a \emph{different} generative family than the audited target model is essential. To probe this, we rerun \ours{} with a \emph{same-family} generator instead, i.e., a generator drawn from the same generative family as the audited target: a VAR paraphraser for RAR-XXL and a Stable Diffusion paraphraser for DiT-RF-XL. Figure~\ref{fig:samefam_ablation} shows that estimation quality drops noticeably in both cases: the same-family $\hat p$ curves no longer track the diagonal and fall well below the cross-family curves at most ground-truth ratios $p$. This is consistent with the argument in Section~\ref{subsec:synt_hold_out_generation}: same-family paraphrases lie near a fixed point of the audited model's own loss and therefore look member-like to its MIA features, leaving the MPE estimator with little signal to separate true members from synthetic non-members.

\paragraph{Scaling with suspect set size.}
We study how estimation accuracy depends on the suspect-set size $|U|$, varying $|U| \in \{100, 500, 1000, 2000\}$ across all evaluated models and MPE estimators (Table~\ref{tab:suspect_size}). The largest improvement occurs when increasing $|U|$ from 100 to 500--1000 samples, after which gains begin to saturate. The trend is consistent across MPE estimators and model families. Notably, even with a tiny suspect set of 100 images, the best-performing estimator already produces a precise estimate (e.g., TIcE on RAR-XXL reaches MAE 0.097), indicating that the method remains practical when only a small amount of suspect data is available.

\begin{table}[h!]
\centering
\small
\renewcommand{\arraystretch}{0.9}
\setlength{\tabcolsep}{2pt}
\vspace{-0.3cm}
\caption{\textbf{Impact of suspect set size on estimation quality.} For each MPE method (AlphaMax, TIcE, PUL-LBE, PUL-NTC-$\tau$MI) we report the MAE of the estimated member ratio $\hat p$ at suspect set sizes of 100, 500, 1{,}000, and 2{,}000 images, averaged over the same set of ground-truth $p$ values as Table~\ref{tab:main}. Estimation quality improves quickly as the suspect set grows, and even at very small sizes ($|U|=100$) several methods (e.g., TIcE on RAR-XXL with MAE $0.097$) already yield useful estimates, indicating that \ours{} remains practical when only a small number of suspect samples is available. Values are reported with standard deviation across 3 runs; lower is better.}
\vspace{0.1cm}
\label{tab:suspect_size}

\begin{tabular}{lccccc}
\toprule
\textbf{Suspect set size} & RAR-XL & RAR-XXL & VAR-24 & VAR-30 & DiT-RF-XL \\
\midrule

\multicolumn{6}{l}{\textbf{AlphaMax}} \\
\midrule
100  & 0.224{\tiny$\pm$0.11} & 0.203{\tiny$\pm$0.10} & 0.226{\tiny$\pm$0.10} & 0.214{\tiny$\pm$0.09} & 0.265{\tiny$\pm$0.14} \\
500  & 0.105{\tiny$\pm$0.05} & 0.101{\tiny$\pm$0.09} & 0.145{\tiny$\pm$0.08} & 0.111{\tiny$\pm$0.05} & 0.268{\tiny$\pm$0.15} \\
1000 & 0.081{\tiny$\pm$0.05} & 0.081{\tiny$\pm$0.05} & 0.126{\tiny$\pm$0.09} & 0.099{\tiny$\pm$0.06} & 0.279{\tiny$\pm$0.16} \\
2000 & 0.077{\tiny$\pm$0.06} & 0.072{\tiny$\pm$0.05} & 0.131{\tiny$\pm$0.10} & 0.089{\tiny$\pm$0.05} & 0.303{\tiny$\pm$0.20} \\
\midrule

\multicolumn{6}{l}{\textbf{TIcE}} \\
\midrule
100  & 0.142{\tiny$\pm$0.06} & 0.097{\tiny$\pm$0.03} & 0.180{\tiny$\pm$0.10} & 0.178{\tiny$\pm$0.08} & 0.356{\tiny$\pm$0.24} \\
500  & 0.113{\tiny$\pm$0.04} & 0.083{\tiny$\pm$0.04} & 0.135{\tiny$\pm$0.06} & 0.114{\tiny$\pm$0.06} & 0.245{\tiny$\pm$0.15} \\
1000 & 0.104{\tiny$\pm$0.05} & 0.083{\tiny$\pm$0.03} & 0.118{\tiny$\pm$0.06} & 0.098{\tiny$\pm$0.05} & 0.223{\tiny$\pm$0.08} \\
2000 & 0.120{\tiny$\pm$0.05} & 0.097{\tiny$\pm$0.03} & 0.136{\tiny$\pm$0.07} & 0.100{\tiny$\pm$0.05} & 0.203{\tiny$\pm$0.05} \\
\midrule

\multicolumn{6}{l}{\textbf{PUL (LBE)}} \\
\midrule
100  & 0.158{\tiny$\pm$0.12} & 0.141{\tiny$\pm$0.12} & 0.167{\tiny$\pm$0.12} & 0.135{\tiny$\pm$0.11} & 0.475{\tiny$\pm$0.25} \\
500  & 0.118{\tiny$\pm$0.09} & 0.092{\tiny$\pm$0.08} & 0.138{\tiny$\pm$0.09} & 0.105{\tiny$\pm$0.06} & 0.248{\tiny$\pm$0.09} \\
1000 & 0.108{\tiny$\pm$0.08} & 0.078{\tiny$\pm$0.08} & 0.126{\tiny$\pm$0.08} & 0.101{\tiny$\pm$0.05} & 0.270{\tiny$\pm$0.11} \\
2000 & 0.103{\tiny$\pm$0.08} & 0.070{\tiny$\pm$0.07} & 0.129{\tiny$\pm$0.08} & 0.094{\tiny$\pm$0.05} & 0.327{\tiny$\pm$0.19} \\
\midrule

\multicolumn{6}{l}{\textbf{PUL (NTC-$\tau$MI)}} \\
\midrule
100  & 0.133{\tiny$\pm$0.10} & 0.142{\tiny$\pm$0.11} & 0.171{\tiny$\pm$0.11} & 0.138{\tiny$\pm$0.07} & 0.217{\tiny$\pm$0.10} \\
500  & 0.104{\tiny$\pm$0.07} & 0.082{\tiny$\pm$0.07} & 0.127{\tiny$\pm$0.07} & 0.091{\tiny$\pm$0.05} & 0.222{\tiny$\pm$0.13} \\
1000 & 0.096{\tiny$\pm$0.06} & 0.058{\tiny$\pm$0.04} & 0.112{\tiny$\pm$0.07} & 0.097{\tiny$\pm$0.05} & 0.219{\tiny$\pm$0.13} \\
2000 & 0.095{\tiny$\pm$0.06} & 0.062{\tiny$\pm$0.05} & 0.127{\tiny$\pm$0.08} & 0.090{\tiny$\pm$0.05} & 0.226{\tiny$\pm$0.14} \\
\bottomrule
\end{tabular}%
\vspace{-0.3cm}
\end{table}

\section{Limitations}
The accuracy of \ours{} depends on two main factors. First, it relies on the strength of available membership-inference signals: when the target model leaks little membership information, MIA features become less separable and our estimates less stable. Thus, \ours{} inherits the limits of current MIA methods, while benefiting from future advances in the field. Second, \ours{} requires a sufficiently strong cross-family generative model to produce synthetic non-members. In domains where no such generator is available, the method does not directly apply.

\section{Conclusions}
We introduced \ourslong{} (\ours{}), a practical framework for dataset usage inference that removes two major barriers in prior work: expensive shadow models and trusted held-out non-member sets. \ours{} generates synthetic non-members via image-to-image paraphrasing, autoencodes the suspect set to reduce distribution shift, extracts diverse MIA features, and formulates dataset usage inference as mixture proportion estimation. Across large image autoregressive and diffusion models, \ours{} accurately estimates member ratios at orders-of-magnitude lower compute than shadow-model-based approaches, requiring only the dataset under examination. We view \ours{} as a practical tool for data owners and auditors seeking quantitative evidence of training-data usage in modern generative models.

\section*{Acknowledgements}

This research was supported by the Polish National Science Centre (NCN) within grants no. 2025/57/N/ST6/04025 and 2023/51/I/ST6/02854.
We gratefully acknowledge Polish high-performance computing infrastructure PLGrid, HPC Center: ACK Cyfronet AGH, for providing computer facilities and support within computational grant no. PLG/2024/017781.
This work was also supported by the German Research Foundation (DFG) within the framework of the Weave Programme under the project titled ``Protecting Creativity: On the Way to Safe Generative Models'', project number 545047250.
We also gratefully acknowledge support from the Initiative and Networking Fund of the Helmholtz Association in the framework of the Helmholtz AI project call under the name ``PAFMIM'', funding number ZT-I-PF-5-227. Responsibility for the content of this publication lies with the authors.

{
    \small
    \bibliographystyle{ieeenat_fullname}
    \bibliography{main.bib}
}

\appendix

\clearpage

\section*{Broader Impact}
\label{app:broader_impact}

This work can help data owners and auditors estimate whether, and to what extent, a particular dataset was used to train a generative model, supporting transparency and accountability around large-scale training data. The same capability, however, may also be misused against the owners of the data: an adversary could in principle use \ours{} as a privacy attack to infer whether someone's private dataset was incorporated into a model's training set without consent, or to extract dataset-level information about confidential training corpora that the model owner has not disclosed.

\section*{Appendix Overview}

The appendix provides additional background and implementation details that support the main paper. Section~\ref{app:iar} reviews image autoregressive models. Section~\ref{app:construct_synth} expands on the synthetic-reference construction used by \ours{}. Section~\ref{app:mias} details the MIA feature suites for autoregressive and diffusion targets. Section~\ref{app:pul_mpe} describes the PUL-based MPE estimator. Section~\ref{app:strength} reports the ablation on the img2img noise strength. We close with broader-impact considerations in Section~\ref{app:broader_impact}.

\section{Image Autoregressive Models}
\label{app:iar}

\paragraph{Notation.}
We denote image dimensions by $C, H, W$, and write $N_t$ for the number of discrete tokens obtained after quantization. An input image is represented as $x \in \mathbb{R}^{C \times H \times W}$, and its autoregressively generated counterpart as $\hat x$. After discretization, the image is mapped to a token sequence $t \in \mathbb{N}^{N_t}$, with $\hat t$ denoting the generated sequence.

\paragraph{Autoregressive modeling for images.}
Classical image autoregressive (AR) models~\citep{chen2020generative} model the joint distribution of an image by predicting tokens sequentially, e.g., in raster-scan order. Letting $t_n$ denote the $n$-th token, the model factorizes
\begin{equation}
p_M(x) = \prod_{n=1}^{N_t} p_M(t_n \mid t_{1:n-1}),
\end{equation}
where $p_M$ is the next-token distribution produced by the target model $M$, and is trained by minimizing the negative log-likelihood
$L_{\text{AR}} = \mathbb{E}_{x \sim \mathcal{D}_{\text{train}}}\left[-\log p_M(x)\right]$. Modeling images at the pixel level produces very long sequences with strong local redundancy and is computationally prohibitive at scale. Methods such as VQ-VAE~\citep{oord2018neuraldiscreterepresentationlearning} and VQ-GAN~\citep{esser2020taming} alleviate this by introducing a discrete latent representation: an encoder compresses the image into a lower-resolution latent grid that is quantized using a learned codebook, the resulting discrete codes form a much shorter token sequence, and a decoder reconstructs the image from these tokens. This token-based formulation aligns image generation with the modeling paradigm used in NLP, so modern image autoregressive models commonly adopt transformer architectures~\citep{vaswani2017attention} similar to those used in GPT-style language models~\citep{radford2019language}.

\paragraph{Visual Autoregressive Models (VAR).}
VAR~\citep{var_tian2024visualautoregressivemodelingscalable} departs from the standard practice of predicting a 1D token sequence in raster order. It instead adopts a multi-scale, coarse-to-fine generation strategy in which images are represented as hierarchical 2D token maps. The model first generates coarse-level tokens and then refines them at progressively higher resolutions, which preserves spatial structure, substantially improves scalability and memory efficiency, and accelerates sampling. VAR exhibits clear scaling behavior analogous to large language models.

\paragraph{Randomized Autoregressive Models (RAR).}
RAR~\citep{rar_yu2024randomizedautoregressivevisualgeneration} addresses a central limitation of standard AR training: its strictly left-to-right conditioning. Inspired by bidirectional objectives such as BERT~\citep{devlin2019bertpretrainingdeepbidirectional}, RAR introduces random permutations of token order during training, allowing the model to learn from a diverse set of bidirectional contexts while still optimizing a likelihood-based AR loss. The permutation schedule gradually anneals toward a fixed raster order, ensuring compatibility with standard autoregressive sampling at inference time, and the resulting bidirectional context yields improved image fidelity.

\section{Synthetic Reference Construction}
\label{app:construct_synth}

Here, we give details for constructing s the synthetic-reference-set used by \ours{}. Given the suspect set $X^{\text{real}}$, the procedure produces both the synthetic non-member set $X^{\text{img2img}}_{\text{nonmember}}$ and the autoencoded suspect set $X^{\text{AE}}$, which serve as the negative reference $N$ and the unlabeled mixture $U$ in Algorithm~\ref{alg:global}, respectively.

Each suspect image $x \in X^{\text{real}}$ is processed twice using components of the same image-to-image generator $\mathcal{G}$. First, it is routed through the autoencoder $(\mathcal{E}, \mathcal{D})$ alone to produce
\[
x^{\text{AE}} = \mathcal{D}(\mathcal{E}(x)).
\]
Second, it is passed through the full image-to-image generator to produce a class-conditioned paraphrase
\[
x^{\text{img2img}}_{\text{nonmember}} = \mathcal{G}(x, p_c;\, s, g).
\]
Routing both sides through the same autoencoder ensures that the unlabeled and negative sets share matching generator-induced artifacts, so that the residual gap between $U$ and $N$ is dominated by membership rather than by low-level distribution shift.

The generator $\mathcal{G}$ is chosen from a different generative family than the audited target model. We use Stable Diffusion~\citep{rombach2022high} for image autoregressive targets and a VAR generator~\citep{var_tian2024visualautoregressivemodelingscalable} for diffusion targets. This cross-family choice avoids the fixed-point artifact described in Section~\ref{subsec:synt_hold_out_generation}: if synthetic non-members are generated by the same family as the audited target, they can become artificially easy or hard under that target's own training loss for reasons unrelated to membership. The class prompt $p_c$ is the ImageNet class label of $x$; the noise strength $s$ and guidance scale $g$ are fixed across all suspect images.

\section{MIA Feature Suite}
\label{app:mias}

This section details the per-sample MIA features that constitute the feature extractor $\Phi_{\text{MIA}}$ defined in Section~\ref{subsec:mia_features}. Recall that
\begin{equation}
\Phi_{\text{MIA}}(x) = \left[f_1(x), \ldots, f_{d_M}(x)\right] \in \mathbb{R}^{d_M},
\end{equation}
where each scalar feature $f_m(x)$ is a membership score obtained by querying the target model $M$ on $x$. Each subsection below specifies which scalar scores $f_m$ enter $\Phi_{\text{MIA}}$ for a given target-model family. We feed the resulting vectors directly into the MPE estimator (Algorithm~\ref{alg:pul-mpe}) and never threshold any individual $f_m$.

\subsection{Image Autoregressive Targets}
\label{app:mias_iar}

For image autoregressive target models we use the suite of \citet{kowalczuk2025privacy}, who develop the first MIAs for IARs by adapting token-level signals from LLMs. Because IARs produce both conditional and unconditional next-token distributions through classifier-free guidance, every base score $\mathcal{M}$ is computed twice (with and without the class condition $c$) and combined into a single feature
\begin{equation}
\Delta \mathcal{M}(x)
= \mathcal{M}\!\left(M(x \mid c)\right) - \mathcal{M}\!\left(M(x \mid \varnothing)\right),
\label{eq:delta_score}
\end{equation}
which sharpens the contrast between members and non-members. Each $\Delta \mathcal{M}$ contributes one scalar feature $f_m$ to $\Phi_{\text{MIA}}(x)$. We list the IAR-specific base scores $\mathcal{M}$ below.

\paragraph{Loss-based score.}
The simplest signal is the per-sample negative log-likelihood under $M$, used in conditional and unconditional modes and combined via Eq.~\ref{eq:delta_score}.

\paragraph{Min-K\%~\citep{shi2024detecting}.}
Min-K\% averages the log-probability of the least likely $K\%$ of tokens in $x$:
\begin{equation}
\mathcal{S}_{\text{Min-K\%}}(x) = \frac{1}{|S_K|} \sum_{n \in S_K} \log p_M(t_n \mid t_{<n}),
\end{equation}
where $S_K$ is the set of token positions with the lowest conditional probabilities. We compute the score in both classifier-free-guidance modes, take their difference, and select the best $K \in \{10, 20, 30, 40, 50\}$.

\paragraph{Min-K\%++.}
Min-K\%++ standardizes the per-token log-probabilities before averaging the bottom $K\%$:
\begin{equation}
\mathcal{S}_{\text{Min-K\%++}}(x)
= \frac{1}{|S_K|}
\sum_{n \in S_K}
\frac{\log p_M(t_n \mid t_{<n}) - \mu_{t_{<n}}}{\sigma_{t_{<n}}},
\end{equation}
where $\mu_{t_{<n}}$ and $\sigma_{t_{<n}}$ are the mean and standard deviation of the next-token distribution at position $n$. The score is again computed in both modes and the difference is used.

\paragraph{Zlib ratio~\citep{zlib2004}.}
The Zlib feature normalizes the model's perplexity by a content-dependent compressibility term:
\begin{equation}
\mathcal{S}_{\text{Zlib}}(x) = \frac{\mathcal{P}_M(x)}{\mathrm{zlib}(x)},
\end{equation}
where $\mathrm{zlib}(x)$ is the byte length of the zlib-compressed input. The numerator is computed in conditional and unconditional modes; the denominator is image-only and shared across both.

\paragraph{CAMIA~\citep{chang2024context}.}
CAMIA extracts several context-aware statistics from the per-token loss sequence---slope, approximate entropy, Lempel--Ziv complexity, the fraction of low-loss tokens, and the loss reduction when the input is repeated. Each statistic is computed under conditional and unconditional decoding and contributes one feature via Eq.~\ref{eq:delta_score}.

\paragraph{SURP~\citep{zhang2024adaptive}.}
SURP averages the next-token probability over tokens that are simultaneously low-probability and emitted under low predictive entropy:
\begin{equation}
\mathcal{S}_{\text{SURP}}(x) = \frac{1}{|\mathcal{S}|} \sum_{n \in \mathcal{S}} p_M(t_n \mid t_{<n}),
\quad
\mathcal{S} = \{n \mid H_n < \epsilon_e,\; p_M(t_n \mid t_{<n}) < \tau_k\},
\end{equation}
where $\tau_k$ is the bottom $k\%$ probability threshold and $H_n$ is the entropy of the predictive distribution at position $n$. We sweep $k \in \{10, 20, 30, 40, 50\}$ and $\epsilon_e \in \{2, 4, 8, 16\}$, and follow Eq.~\ref{eq:delta_score} for the final feature.

\subsection{Diffusion Targets}
\label{app:mias_diffusion}

For diffusion target models we adopt the feature suite of CDI~\citep{dubinski2025cdi}, which aggregates several gray-box membership signals computed from the model's noise prediction $M(z_t, t)$ on the latent $z = \mathcal{E}(x)$ at carefully chosen timesteps. We use the same encoder $\mathcal{E}$ as in Section~\ref{subsec:synt_hold_out_generation}; $\bar\alpha_t$ denotes the standard cumulative noise schedule. The descriptions below summarize the formulations from \citet{dubinski2025cdi} and the underlying papers; we refer the reader to those works for full derivations. Each scalar score below contributes one feature $f_m$ to $\Phi_{\text{MIA}}(x)$.

\paragraph{Denoising loss~\citep{carlini2023extracting_diffusion}.}
Following the loss-attack tradition for discriminative models~\citep{yeom2018privacy}, this score evaluates the squared denoising error at a fixed timestep $t^{*}{=}100$:
\begin{equation}
\mathcal{S}_{\text{Loss}}(x) = \frac{1}{K} \sum_{k=1}^{K} \left\| \epsilon_k - M\!\left(z_{t^{*}}^{(k)}, t^{*}\right) \right\|_2^{2},
\quad z_{t^{*}}^{(k)} = \sqrt{\bar\alpha_{t^{*}}}\, z + \sqrt{1 - \bar\alpha_{t^{*}}}\, \epsilon_k,
\end{equation}
averaged over $K{=}5$ independent noise samples $\epsilon_k \sim \mathcal{N}(0, I)$ to reduce variance. Members are expected to receive lower loss than non-members.

\paragraph{SecMI$_{\text{stat}}$~\citep{duan23bSecMI}.}
SecMI$_{\text{stat}}$ measures how well a deterministic DDIM~\citep{song2020denoising} denoising step inverts the corresponding sampling step, an operation the target model is expected to perform more accurately on members. Letting $\Phi_M(z, t)$ denote the deterministic DDIM reverse trajectory from $z$ to timestep $t$, $\phi_M(\cdot, t)$ the DDIM sampling-inverse step, and $\psi_M(\cdot, t)$ the DDIM denoising step, the score is the $t$-error
\begin{equation}
\mathcal{S}_{\text{SecMI}}(x) = \left\| \psi_M\!\left(\phi_M\!\left(\Phi_M(z, t^{*}), t^{*}\right), t^{*}\right) - \Phi_M(z, t^{*}) \right\|_2^{2},
\end{equation}
with $t^{*}{=}100$. Intuitively, this is the round-trip error of one DDIM step starting from the deterministic reverse latent $\Phi_M(z, t^{*})$; members are expected to yield a lower $t$-error.

\paragraph{PIA~\citep{kong2024an}.}
The Proximal Initialization Attack (PIA) compares the noise prediction on a clean latent with the noise prediction on a noised latent constructed from it: a model that has memorized $x$ will track this trajectory more closely. The score is
\begin{equation}
\mathcal{S}_{\text{PIA}}(x) = \left\| M(z, 0) - M\!\left( \sqrt{\bar\alpha_{t^{*}}}\, z + \sqrt{1 - \bar\alpha_{t^{*}}}\, M(z, 0),\; t^{*} \right) \right\|_p,
\end{equation}
with $t^{*}{=}200$ and $p{=}5$. Members are expected to receive lower scores.

\paragraph{PIAN~\citep{kong2024an}.}
PIAN is a normalized variant of PIA that rescales the clean-latent noise prediction so that it more closely matches the standard-Gaussian assumption on $M(z, 0)$:
\begin{equation}
\hat M(z, 0) = \frac{C \cdot H \cdot W}{\sqrt{\pi/2}} \cdot \frac{M(z, 0)}{\| M(z, 0) \|_1}.
\end{equation}
The PIA score is then computed as above with $\hat M(z, 0)$ in place of $M(z, 0)$.

\paragraph{CLiD~\citep{clid}.}
The Conditional Likelihood Discrepancy attack (CLiD) exploits classifier-free-guidance models by contrasting the conditional and unconditional denoising losses at the same timestep:
\begin{equation}
\mathcal{S}_{\text{CLiD}}(x)
= \left\| \epsilon - M(z_{t^{*}}, t^{*} \mid c) \right\|_2^{2}
- \left\| \epsilon - M(z_{t^{*}}, t^{*} \mid \varnothing) \right\|_2^{2},
\end{equation}
which acts as a diffusion analog of the $\Delta$ score in Eq.~\ref{eq:delta_score} and is more discriminative than either branch alone for class-conditional and text-conditional models~\citep{dubinski2025cdi}. The rest of our pipeline is identical to the IAR case in Section~\ref{subsec:MPE_method}.

\section{PUL-based Mixture Proportion Estimation}
\label{app:pul_mpe}

This section details the MPE construction that underlies the PUL-LBE and PUL-NTC-$\tau$MI rows of Table~\ref{tab:main}. It instantiates the abstract $\textsc{MPE}(Z^{\text{MIA}}, U, N)$ call from Algorithm~\ref{alg:global}, where $U = X^{\text{AE}}$ is the unlabeled set, $N = X^{\text{img2img}}_{\text{nonmember}}$ the labeled-negative set, and $Z^{\text{MIA}} = \Phi_{\text{MIA}}(U \cup N)$ the corresponding MIA features. The construction has three steps: (i) train a Positive--Unlabeled classifier on $(U, N)$, (ii) project both sets onto its scalar score $s(x) \in \mathbb{R}$, and (iii) estimate the member proportion $\pi$ of $U$ by histogram matching on this score, returning $\hat p$ as our estimate of $\pi$.

\paragraph{PUL classifiers.}
We use two distinct paradigms for modeling labeled and unlabeled data. \textbf{LBE}~\citep{gong2021lbe} is a probabilistic, instance-dependent PU learner that assumes the labeling probability depends on both features $\mathbf{x}$ and latent labels $y$, and factorizes the joint as
\begin{equation}
P(y, \ell \mid \mathbf{x}) = P(y \mid \mathbf{x})\, P(\ell \mid y, \mathbf{x}),
\end{equation}
where $\ell \in \{0, 1\}$ indicates whether a sample is observed as labeled, i.e., whether it belongs to $N$. The class-posterior parameters $\theta_1$ and the labeling-mechanism parameters $\theta_2$ are jointly fit by maximizing the marginal likelihood
\begin{equation}
\max_{\theta = (\theta_1, \theta_2)} \prod_{i=1}^n P(\ell_i \mid \mathbf{x}_i; \theta) = \max_{\theta} \prod_{i=1}^n \sum_{y_i \in \{0,1\}} P(\ell_i, y_i \mid \mathbf{x}_i; \theta)
\end{equation}
via the Expectation--Maximization (EM) algorithm.
\textbf{NTC-$\tau$MI}~\citep{teser2025threshold} instead trains a Non-Traditional Classifier (NTC) on $(U, N)$ and then chooses a decision threshold $\tau$ that maximizes the mutual information between the binarized predictions $\hat Y^{\tau}$ and the continuous NTC scores $Z$,
\begin{equation}
\tau_{*} = \arg\max_{\tau}\; MI(\hat Y^{\tau}, Z),
\end{equation}
estimating the mixed-type $MI$ via a $k$-nearest-neighbors estimator~\citep{ross2014knnmi} that runs in $O(n \log n)$ time on the one-dimensional score $Z$.

\begin{figure}[t]
    \centering
    \includegraphics[width=0.9\linewidth]{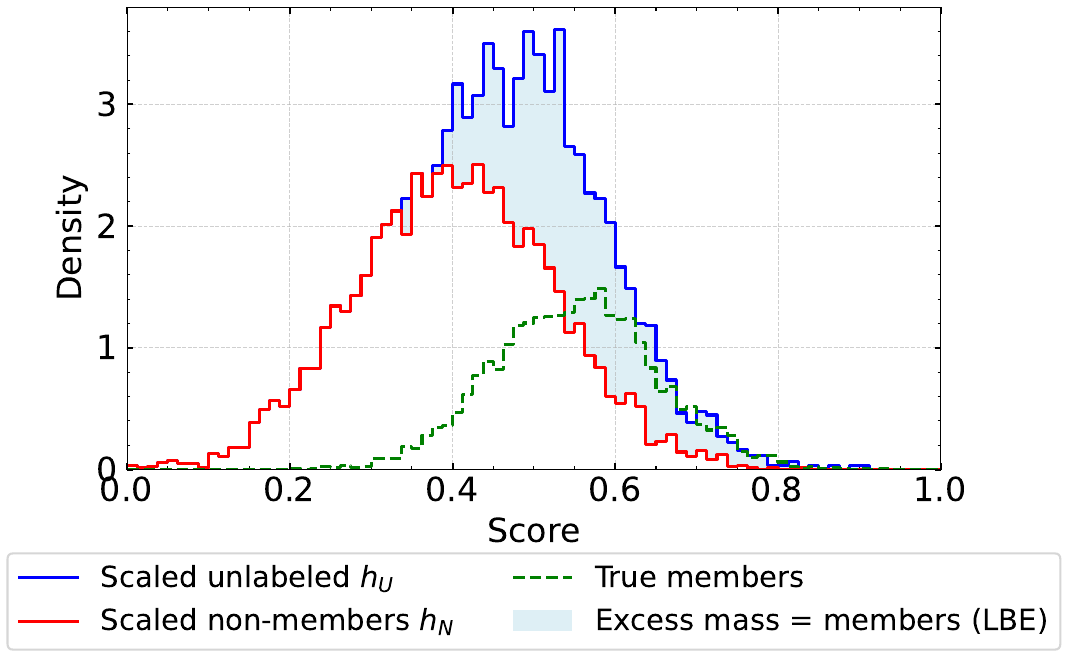}
    \caption{\textbf{Density of membership scores produced by a PUL classifier.} The unlabeled set $U$ is treated as a mixture of positive and negative samples; $h_N$ and $h_U$ are the score histograms of the \textcolor{red}{known synthetic negatives} ($N$) and the \textcolor{blue}{unlabeled} set ($U$), respectively. In each bin, the portion of $h_U$ that lies above $h_N$ reflects \textcolor{lightblue}{positives}, i.e., the ``excess mass.'' Integrating this excess across all bins recovers the estimated fraction of positive samples $\hat p$ in $U$.}
    \label{fig:histogram_estimation}
\end{figure}

\paragraph{Histogram-based ratio estimation.}
After training, both methods reduce to a one-dimensional score $s(x) \in \mathbb{R}$ for each $x \in U \cup N$. We bin these scores into a fixed grid $\mathcal{B}$ and form histograms $h_U(s)$ and $h_N(s)$, treating $h_U$ as an empirical estimate of the score-space mixture $p_U(s) = \pi\, p_M(s) + (1-\pi)\, p_N(s)$ from Section~\ref{subsec:MPE_method}. We then solve for the largest non-member-mass coefficient $\hat\pi_N$ that allows $h_U$ to be approximately explained by a scaled $h_N$:
\begin{equation}
\hat\pi_N = \max\!\left\{ \pi_N \in [0, 1] \;\middle|\; h_U(s) - \pi_N\, h_N(s) \ge -\delta,\ \forall s \in \mathcal{B} \right\},
\end{equation}
where $\delta > 0$ is a small relaxation that absorbs finite-sample noise (see Figure~\ref{fig:histogram_estimation}). All histograms share identical bin edges and extremely sparse bins are masked. The estimated member ratio is then $\hat p = 1 - \hat\pi_N$, our MPE estimate of $\pi$. We rerun the full pipeline with $3$ independent random seeds and report the mean and standard deviation of $\hat p$ in Tables~\ref{tab:main} and~\ref{tab:suspect_size}.

\begin{figure}[h!]
\centering
\begin{minipage}{0.8\linewidth}
\hrule height 1.5pt
\vspace{2pt}
\captionof{algorithm}{\textit{PUL-based MPE} (\textsc{MPE} subroutine of Algorithm~\ref{alg:global})}
\label{alg:pul-mpe}
\vspace{2pt}
\hrule
\begin{algorithmic}[1]
\small
\Require MIA features $Z^{\text{MIA}}$; unlabeled set $U$; synthetic negatives $N$; bin edges $\mathcal{B}$; relaxation $\delta$
\State $\ell(x) \gets \mathbb{1}[x \in N]$ for all $x \in U \cup N$ \Comment{observed-label indicator}
\State Train PUL classifier on $(Z^{\text{MIA}}, \ell)$
\State Compute classifier score $s(x)$ for all $x \in U \cup N$
\State $h_U \gets \textsc{Hist}(\{s(x) : x \in U\},\, \mathcal{B})$,\quad $h_N \gets \textsc{Hist}(\{s(x) : x \in N\},\, \mathcal{B})$
\State $\hat\pi_N \gets \max\{\pi_N \in [0, 1] : h_U(s) - \pi_N\, h_N(s) \ge -\delta,\ \forall s \in \mathcal{B}\}$
\State $\hat p \gets 1 - \hat\pi_N$
\State \Return $\hat p$
\end{algorithmic}
\vspace{2pt}
\hrule height 1.5pt
\end{minipage}
\end{figure}

\paragraph{Assumptions.}
The estimator $\hat p$ is consistent and asymptotically unbiased under four standard MPE conditions:
\begin{enumerate}
    \item \textit{Mixture assumption.} The unlabeled score distribution is a convex mixture $p_U(s) = \pi\, p_M(s) + (1-\pi)\, p_N(s)$ of the member and non-member score distributions, as in Section~\ref{subsec:MPE_method}.
    \item \textit{Representative non-members.} The reference set $N$ faithfully represents the non-member component of $U$, with no covariate or domain shift between $N$ and the non-member portion of $U$.
    \item \textit{Support separation.} There exists at least one region in score space where members occur but non-members do not, ensuring identifiability of $\pi$.
    \item \textit{Sufficient sample size.} The histograms $h_U$ and $h_N$ approximate the true densities closely enough that sampling noise does not dominate the mixture-ratio constraint.
\end{enumerate}
When these conditions hold, $\hat p$ converges to the true member ratio $\pi$ as the sample sizes of $U$ and $N$ increase. Our framework is not tied to any specific PUL algorithm; other estimators from the literature can be plugged in in place of LBE or NTC-$\tau$MI~\citep{bekker2019sarem, gerych2022recovering, kiryo2017pulnonnegative}.

\section{Additional Experiments}
\label{app:additional_experiments}

\subsection{Effect of the img2img Noise Strength $s$}
\label{app:strength}

\paragraph{Setup.}
Algorithm~\ref{alg:construct_synth} constructs the synthetic non-member set $X^{\text{img2img}}_{\text{nonmember}}$ with the Stable Diffusion img2img pipeline, where the noise strength $s \in [0, 1]$ controls how much of the original latent is replaced by Gaussian noise before denoising. Throughout the main paper we fix $s = 0.5$. Two opposing pressures shape this choice. For $s$ \emph{too small}, the paraphrase $\mathcal{G}(x, p_c;\, s, g)$ stays close to the autoencoder reconstruction $\mathcal{D}(\mathcal{E}(x))$ used to build $X^{\text{AE}}$, so the unlabeled set $U$ and the synthetic negatives $N$ collapse onto each other in the MIA feature space and the MPE estimator sees little residual signal between members and non-members. For $s$ \emph{too large}, the paraphrase departs significantly from the original semantics, and the dominant gap between $U$ and $N$ becomes a generic content shift rather than a membership signal---the same failure mode that drives the \textit{Synth} rows of Table~\ref{tab:main} away from the \textit{Real} oracle. We therefore expect a sweet spot at intermediate $s$, and we ablate this choice empirically by re-running the full \ours{} pipeline (\textit{Synth + AE} configuration only) at $s \in \{0.4, 0.5, 0.6\}$ for the two image autoregressive targets VAR-24 and RAR-XL, keeping every other component unchanged. For these two targets, the $s = 0.5$ column of Table~\ref{tab:strength_ablation} reproduces the \textit{Synth + AE} block of Table~\ref{tab:main}.

\begin{table}[h!]
\centering
\small
\renewcommand{\arraystretch}{1}
\setlength{\tabcolsep}{4pt}
\caption{\textbf{Ablation on the img2img noise strength $s$ in the \textit{Synth + AE} configuration of \ours{}} (suspect set size 1{,}000 images; mean and max absolute error over ground-truth ratios $p \in \{0.1, \ldots, 1.0\}$). For each target model and MPE estimator we report \textit{MAE} and \textit{max MAE} of the estimated member ratio $\hat p$ at three noise strengths used in Algorithm~\ref{alg:construct_synth}. \textbf{Bold} marks the strength with the lowest MAE in each row. The $s = 0.5$ column reproduces the \textit{Synth + AE} block of Table~\ref{tab:main} for these two targets. Across the 8 (target, estimator) cells, $s = 0.5$ achieves the lowest mean MAE and the lowest mean max MAE; $s = 0.4$ is uniformly the worst, while $s = 0.6$ is comparable to $s = 0.5$ in mean MAE but markedly less stable in the worst case. Lower is better.}
\vspace{0.1cm}
\label{tab:strength_ablation}
\begin{tabular}{l l cc cc cc}
\toprule
& & \multicolumn{2}{c}{$s = 0.4$} & \multicolumn{2}{c}{$s = 0.5$} & \multicolumn{2}{c}{$s = 0.6$} \\
\cmidrule(lr){3-4} \cmidrule(lr){5-6} \cmidrule(lr){7-8}
\textbf{Target} & \textbf{Estimator} & \textbf{MAE} & \textbf{max MAE} & \textbf{MAE} & \textbf{max MAE} & \textbf{MAE} & \textbf{max MAE} \\
\midrule
\multirow{4}{*}{VAR-24}
 & TIcE~\citep{bekker2018tice}                   & 0.227 & 0.419 & \textbf{0.118} & 0.232 & 0.122          & 0.215 \\
 & AlphaMax~\citep{jain2016alphamax}             & 0.195 & 0.432 & 0.126          & 0.315 & \textbf{0.115} & 0.310 \\
 & PUL (LBE~\citep{gong2021lbe})                 & 0.196 & 0.339 & \textbf{0.126} & 0.251 & 0.152          & 0.361 \\
 & PUL (NTC-$\tau$MI~\citep{teser2025threshold}) & 0.178 & 0.342 & \textbf{0.112} & 0.220 & 0.140          & 0.309 \\
\midrule
\multirow{4}{*}{RAR-XL}
 & TIcE~\citep{bekker2018tice}                   & 0.317 & 0.555 & 0.104          & 0.162 & \textbf{0.103} & 0.189 \\
 & AlphaMax~\citep{jain2016alphamax}             & 0.222 & 0.416 & \textbf{0.081} & 0.174 & 0.082          & 0.240 \\
 & PUL (LBE~\citep{gong2021lbe})                 & 0.187 & 0.405 & \textbf{0.108} & 0.269 & 0.119          & 0.291 \\
 & PUL (NTC-$\tau$MI~\citep{teser2025threshold}) & 0.235 & 0.431 & 0.096          & 0.209 & \textbf{0.090} & 0.182 \\
\midrule
\multicolumn{2}{l}{\textit{Mean over (target, estimator) cells}} & 0.220 & 0.417 & \textbf{0.109} & \textbf{0.229} & 0.115 & 0.262 \\
\bottomrule
\end{tabular}
\end{table}

\paragraph{$s = 0.4$ is uniformly worse.}
At the lower noise strength the synthetic paraphrase $\mathcal{G}(x, p_c;\, s, g)$ stays close to the autoencoder reconstruction $\mathcal{D}(\mathcal{E}(x))$ used to construct $X^{\text{AE}}$, so the unlabeled mixture $U$ and the synthetic negatives $N$ become barely separable in the MIA feature space and the MPE estimator loses the membership signal. MAE roughly doubles or triples relative to $s = 0.5$ on every (target, estimator) cell (e.g., RAR-XL/TIcE: $0.317$ vs.\ $0.104$; VAR-24/PUL-LBE: $0.196$ vs.\ $0.126$), and the max MAE consistently exceeds $0.3$. Across all 8 cells, the mean MAE at $s = 0.4$ ($0.220$) is more than $2\times$ the mean MAE at $s = 0.5$ ($0.109$).

\paragraph{$s = 0.5$ is the most robust choice.}
$s = 0.5$ attains the lowest MAE in $5/8$ cells and is within $0.011$ of the best in the remaining three. Averaged over the 8 (target, estimator) cells, $s = 0.5$ gives the lowest mean MAE ($0.109$) and the lowest mean max MAE ($0.229$), beating both $s = 0.4$ ($0.220$, $0.417$) and $s = 0.6$ ($0.115$, $0.262$). The improvement over $s = 0.6$ in the worst-case error is the practically relevant metric for an auditing tool, where pathological prevalences should not blow up the estimate: e.g., VAR-24/PUL-LBE max MAE drops from $0.361$ at $s = 0.6$ to $0.251$ at $s = 0.5$, and RAR-XL/AlphaMax from $0.240$ to $0.174$.

\paragraph{Increasing $s$ to $0.6$ does not pay off.}
While $s = 0.6$ marginally outperforms $s = 0.5$ in MAE on $3/8$ cells, the wins are small ($\le 0.011$, with two of the three differences below $0.007$) and are offset by larger losses on the other 5 cells (up to $0.028$ in MAE and $0.110$ in max MAE). Pushing $s$ further toward $1$ would amount to nearly unconditional generation, which Section~\ref{subsec:synt_hold_out_generation} already argues against on conceptual grounds.

\paragraph{Conclusion.}
Our experimental observations support the choice $s = 0.5$ in Algorithm~\ref{alg:construct_synth}: large enough that the img2img paraphrase departs meaningfully from the autoencoded suspect, but not so large that semantic drift dominates over the membership signal. We therefore use $s = 0.5$ throughout the main paper.

\end{document}